\def\year{2020}\relax
\documentclass[letterpaper]{article} 
\usepackage{aaai20}  
\usepackage{times}  
\usepackage{helvet} 
\usepackage{courier}  
\usepackage[hyphens]{url}  
\usepackage{graphicx} 
\usepackage{graphicx}  
\usepackage{amsmath}

\title{Projection Mapping Implementation: Enabling Direct Externalization of Perception Results and Action Intent to Improve Robot Explainability}
\author{
Zhao Han$^{\dagger}$\thanks{$^{\dagger}$Corresponding author: {\tt\small zhan@cs.uml.edu}}, Alexander Wilkinson, Jenna Parrillo, Jordan Allspaw, Holly A. Yanco
\\Department of Computer Science
\\University of Massachusetts Lowell
\\One University Ave, Lowell, MA, USA
}

\makeatletter
\def\thanks#1{\protected@xdef\@thanks{\@thanks
        \protect\footnotetext{#1}}}
\makeatother

\begin{document}

\maketitle

\begin{abstract}
Existing research on non-verbal cues, e.g., eye gaze or arm movement, may not accurately present a robot's internal states such as perception results and action intent. Projecting the states directly onto a robot's operating environment has the advantages of being direct, accurate, and more salient, eliminating mental inference about the robot's intention. However, there is a lack of tools for projection mapping in robotics, compared to established motion planning libraries (e.g., MoveIt). In this paper, we detail the implementation of projection mapping to enable researchers and practitioners to push the boundaries for better interaction between robots and humans. We also provide practical documentation and code for a sample manipulation projection mapping on GitHub: \url{https://github.com/uml-robotics/projection_mapping}.
\end{abstract}
 
\section{Introduction}

As robots are increasingly deployed in areas ranging from factories and warehouses to hotels and private homes, there has been a growing interest in having robots explain their behaviors and actions. Past research has shown that improving the understanding of robots improves trust in real-time \cite{desai2013trust} and leads to greater efficiency during more difficult human-robot collaboration scenarios \cite{admoni2016nonverbal}.

Traditionally, because of the distinct embodiment feature of physical robots, human-robot interaction (HRI) researchers have been focused on how to enable robots to communicate their intention through non-verbal means that are typical among humans, most notably pointing through eye gaze \cite{moon2014meet,admoni2017eyegaze}, and arm movement \cite{dragan2013legibility,kwon2018expressing}. Light as an indication has also been used \cite{szafir2015communicating}.

\begin{figure}[t]
\centering
\includegraphics[width=0.9\columnwidth]{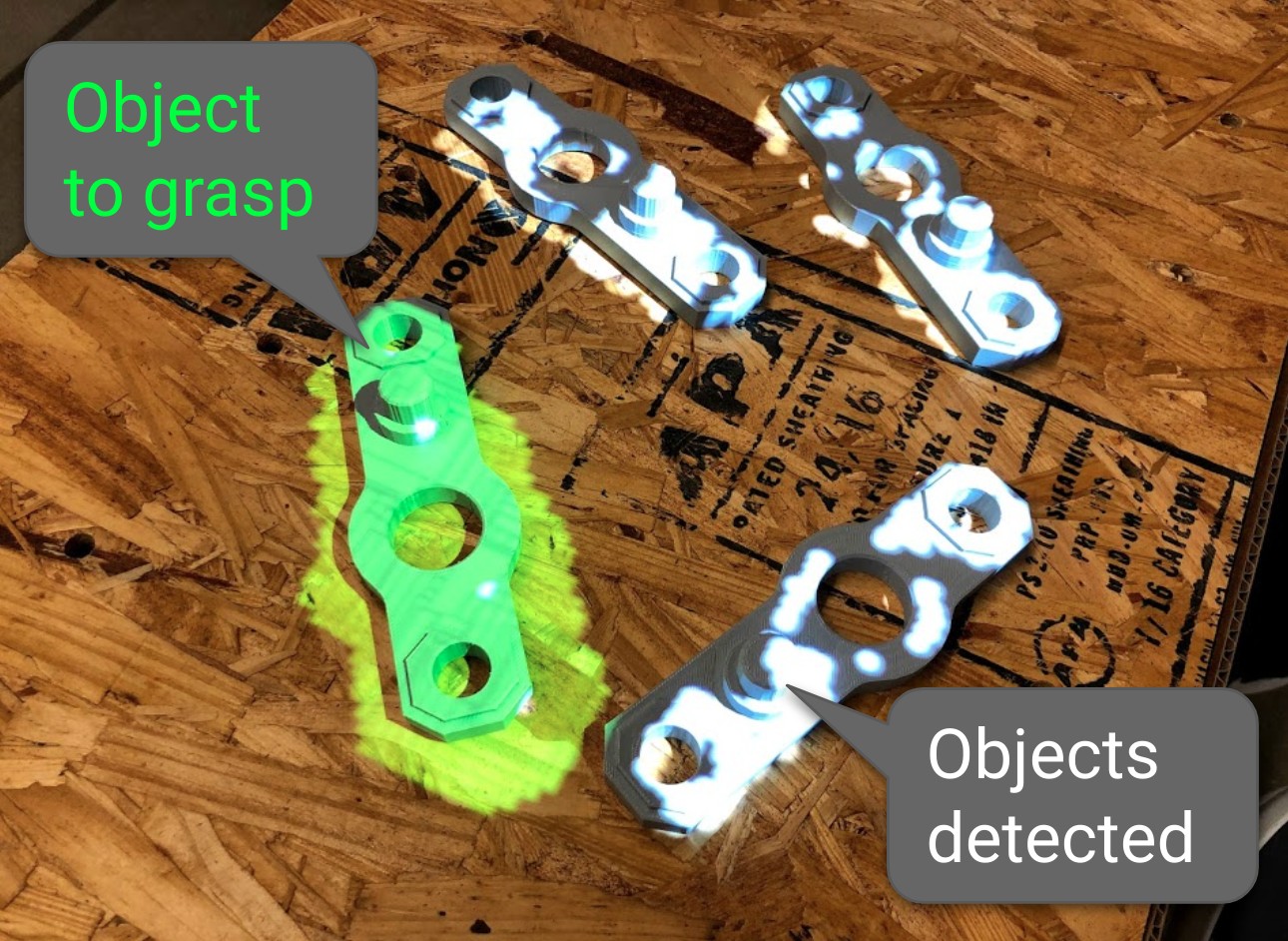} 
\caption{The projection of perception results: the detected objects (white and green) and the object to be manipulated (green). Using our implementation of projection mapping, researchers and practitioners can enable a robot to accurately externalize internal states for explanation. A video is available at \url{https://youtu.be/S0z9e2gUrEA}.}
\label{fig:projection}
\end{figure}

These non-verbal cues found in human life play an important role in supporting and improving communication \cite{admoni2015nvc}, 
but can also cause confusion.
For example, how can a robot communicate which objects it has detected and which one is it going to grasp? In cases like these, using eye-gaze and/or pointing with its arm or end effector can be vague, especially in a clustered environment (e.g., for four clustered objects on a table). Even with verbal explanations, these gestures could be underspecified, requiring follow-up questions for clarification.

In this paper, we present a tool for implementing projection mapping using an off-the-shelf projector, including the high level architecture and low level technical details, in order to project perception results directly onto non-flat objects of interest in the environment. 
We also describe a concrete robot and hardware platform as an example of the tool's use, even though the technique is robot-agnostic. Note that while we focus on projection mapping with the same object as input and output, projection mapping can be applied more broadly to arbitrary objects and targets \cite{grundhofer2018recent}.

\begin{figure*}[t]
\centering
\includegraphics[width=0.9\textwidth]{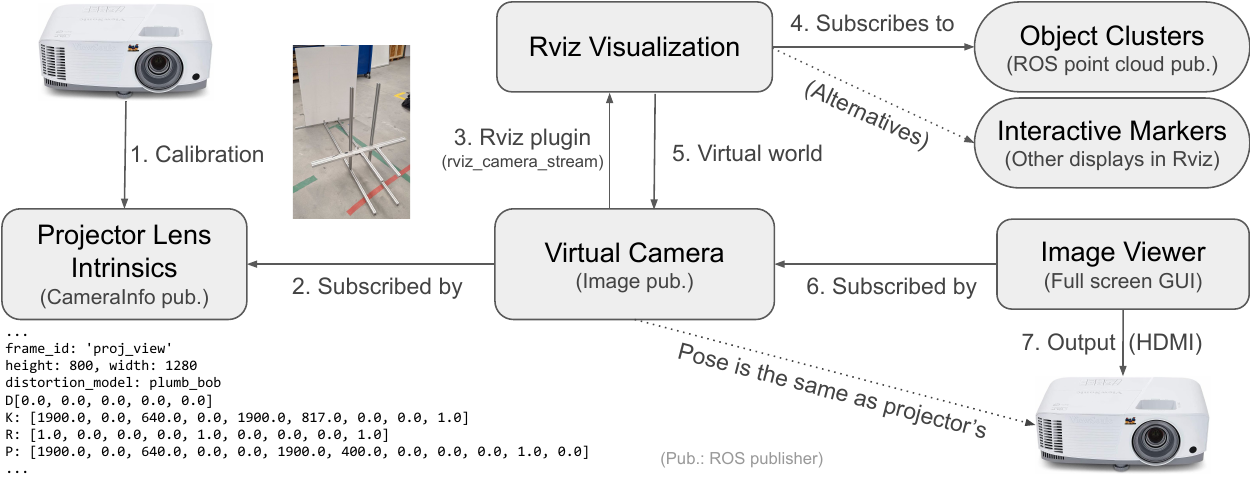} 
\caption{High-level diagram for our projection mapping implementation. With the projector lens calibrated, a virtual camera -- placed in Rviz with the same pose as the projector in real world -- subscribes to the camera intrinsics so it can output an image of objects visualized in the virtual world in Rviz to the projector to reflect the perceived objects. See Section \ref{sec:impl} for more details.}
\label{fig:diagram}
\end{figure*}

While projection mapping in robotics is not a new idea, the implementation effort has been missing and thus not as accessible as implementing arm movement or eye gaze through head movement. This gap effectively blocks HRI researchers from conducting human-subject studies to investigate the effects of accurate externalization through projection mapping or by comparing it to other methods. In addition, this work can also help robotics and AI researchers to externalize the output of their computer vision algorithms for a better understanding of their algorithms.

As opposed to the non-verbal methods and verbal explanations found among humans, projection mapping is a method that allows for \textit{direct and accurate} externalization. This projection completely removes the need for mental inference as the perceived objects or the objects to be manipulated are directly externalized. The directness is similar to the use of a display screen, but projection is more \textit{salient} because bystanders or robot coworkers can also see the projection from farther away, instead of requiring people to stop their work in order to walk to a monitor to examine the robot's states. Direct projection onto the operating environment also \textit{eliminates mental mapping} from another media such as a monitor, which can cause misjudgment and lead to undesired consequences.

\section{Related Work}

Compared to typical use cases of a projector, such as watching videos or presenting slides, the usage of a projector in the robotics field is rather rare. A more popular approach has been leveraging virtual reality headsets (e.g., \cite{labvr,chakraborti2018projection,rosen2020communicating}), which are readily available commercially, including devices such as the Oculus Rift and the HTC Vive along with their Software Development Kits. Some researchers have explored use cases of projection mapping (e.g., \cite{andersen2016projecting}) or simple projection methods (e.g., \cite{chadalavada2015s,watanabe2015communicating,coovert2014spatial,ghiringhelli2014interactive}) that require no extra headset to be worn by the users; however, as previously stated, the lack of readily available implementation has impeded further advancement.

\citeauthor{andersen2016projecting} used a projector to reveal a robot's intent and task information onto the workspace for better human-robot collaboration \shortcite{andersen2016projecting}. Specifically, the robot constantly projects the wireframe of a car door to help a factory worker to understand the robot's perception accuracy; additionally, before part manipulation, the robot will project the segment of interest to inform the worker of its next manipulation target. An experiment was conducted to compare projection with display screens for a cube moving and rotating task. Results show that there were fewer performance errors and fewer questions asked with projection.

Recently, \citeauthor{gao2019pati} proposed \cite{gao2019pati} using a projector to project an interface on a flat tabletop surface for robot programming. However, the focus is on the ease of robot programming, and the projection does not require mapping objects back to their corresponding real-world objects. Similarly, many researchers have been using projection to project onto another flat surface, the floor, to indicate navigation intent. \citeauthor{chadalavada2015s} used lines to indicate the path plan and the collision avoidance range \shortcite{chadalavada2015s}. \citeauthor{watanabe2015communicating} used a band of light to externalize the path \shortcite{watanabe2015communicating} while \citeauthor{coovert2014spatial} used arrows for the path \shortcite{coovert2014spatial}.

In our work, one of the implementation efforts was mapping perceived objects (e.g., from a depth camera) back to another view point (projector) with the same distortion caused by the view point difference, rather than flat surface projection. Nonetheless, we also provide a navigation path projection using our method in the aforementioned GitHub repository.

\section{Projection Mapping Implementation}\label{sec:impl}

Figure \ref{fig:diagram} shows the high-level architecture of our implementation. 
Essentially, we set up a virtual camera with the same lens intrinsics in the same pose as the projector's in the physical world in the ROS visualization software, Rviz. We then publish perception results in point cloud clusters and the object point cloud to be manipulated, then add point cloud visualization in Rviz. Finally, we have an image viewer in a full screen GUI to output the image that the virtual camera sees to a projector. We implemented everything in ROS \cite{quigley2009ros}. All of the files are available on GitHub\footnote{\url{https://github.com/uml-robotics/projection_mapping}}, including a sample Rviz config file, sample point clouds in pcd format, and the launch files for publishing lens intrinsics and the pose of the projector.

Our projection mapping system functions on the principal that a projector is the dual of a camera. If a camera is thought of as a map from 3D world coordinates to 2D image plane coordinates, then a projector is a map from 2D image plane coordinates to 3D world coordinates. In a camera, each light ray from the world passes through the lens and hits the sensor. In a projector, each light ray passes 
through the lens and hits a surface in the world.

By using a virtual camera with the same intrinsics and extrinsics as the projector, we can then map points from the virtual world's 3D space to 3D space in the real world.

\subsection{Projector Selection Consideration}

While any projector should theoretically work as long as we calibrate it to get the lens intrinsics as detailed below, there were a couple of factors that made us choose the ViewSonic PA503W projector in our implementation.
In the common use case where projectors are used to watch movies and show presentation slides, the room lighting is often switched off or dimmed to make the projection more bright and thus more legible. However, robots often operate indoors with lights on and not dimmed, so the consideration here is to make the projection visible and legible even under bright light conditions. 

There are three contributing factors: brightness, contrast and the projection technology. The standard measure of brightness is the ANSI lumens value, which is a measure of the total light output per unit of time. For reference, our PA503W produces 3,800 ANSI lumens of light. Contrast is expressed by a ratio between the darkest and lightest areas of an image, with our PA503W being capable of a 22,000:1 contrast ratio. Finally, there are two popular projection display technologies: Liquid Crystal Display (LCD) and DLP \cite{hornbeck1997digital}. As a brief summary from the work by \citeauthor{hornbeck1997digital}, DLP is based on a Digital Micromirror Device (DMD), and it has a higher brightness than LCD because DMD has a reflective and high-fill factor digital light switch, as opposed to active-matrix LCDs, which are transmissive and thus the generated heat cannot be dispatched well. See \cite{hornbeck1997digital} for more detail.

\subsection{Projector Calibration}\label{sec:calibration}



\begin{figure}[t]
\centering
\includegraphics[width=0.5\columnwidth]{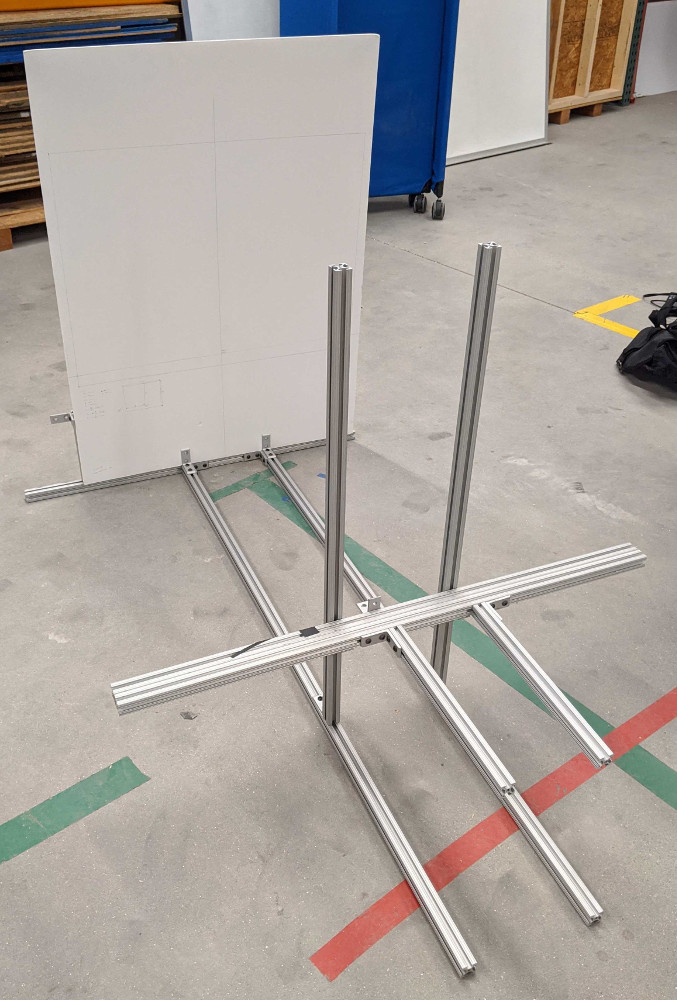} 
\caption{We calculated the intrinsics of our particular projector by mounting it perpendicular to a posterboard at a fixed distance. See Section \ref{sec:calibration} for more details.}
\label{fig:calibration-jig}
\end{figure}

We modelled the projector with a pinhole lens model after determining the focal length $f$ and principal point $(c_x, c_y)$. Both of these values were calculated manually by mounting the projector perpendicular to a flat surface at some fixed distance (e.g., \ref{fig:calibration-jig}), marking the corners of the image, and then using the projective equation:
\begin{equation}
    f = w \frac{Z}{W}
\end{equation}

\noindent where $w$ is the width of the projected image in pixels, $W$ is the width of the projected image in meters, and $Z$ is the distance from the projector to the image plane in meters.

The principal point $(c_x, c_y)$ can be calculated given $X$ and $Y$, the respective distances from the origin to the intersection of the optical plane and the optical axis in meters:


\noindent\begin{minipage}{.5\linewidth}
\begin{equation}
   c_x = w \frac{X}{W}
\end{equation}
\end{minipage}%
\begin{minipage}{.5\linewidth}
\begin{equation}
  c_y = h \frac{Y}{H}
\end{equation}
\end{minipage}

\noindent where $h$ is the height of the projected image in pixels.

These values for $f$, $c_x$, and $c_y$ were then placed into an intrinsic camera matrix $K$:
\begin{equation}
K=
\begin{bmatrix}
	f_x & 0   & c_x \\
	0   & f_y & c_y \\
	0   & 0   & 1
\end{bmatrix}
\end{equation}

\noindent With this matrix calculated for our projector, we then publish it in a ROS CameraInfo message\footnote{\url{https://docs.ros.org/melodic/api/sensor_msgs/html/msg/CameraInfo.html}} and set up a virtual camera in RViz (see Section~\ref{sec:virtual-camera}) that subscribes to the CameraInfo message and is located at a TF transform corresponding to the pose of the physical projector. The CameraInfo message for the ViewSonic projector and the details of this process are in our code\footnote{\url{https://github.com/uml-robotics/projection_mapping/blob/master/projector_camera_info.yaml}}.


It is worth noting that measurement error may accumulate during this manual process. Since consumer projectors are typically designed with little radial or tangential distortion, we did not explicitly model lens distortion. In practice, the projection does not deviate much from the real objects, but it could be improved by more accurate projector-camera calibration methods~\cite{moreno2012simple}.

\subsection{Virtual Camera}\label{sec:virtual-camera}

We developed \texttt{rviz\_camera\_stream}\footnote{\url{https://github.com/uml-robotics/rviz_camera_stream}; thanks to Lucas Walter (\url{https://github.com/lucasw}) for his contribution.}, an Rviz plugin that outputs an image of what a camera sees in the Rviz virtual world. We refer this as the \textit{virtual camera}.

Represented in a ROS TF frame \cite{foote2013tf}, the pose of the virtual camera -- placed in the virtual world in Rviz -- is the same as the pose of the projector in the real-world. Together with the fact that the virtual camera pose is in the same transform hierarchy as a perception sensor, we are able to transform a point cloud or any other visualizations from the perception sensor's frame to the virtual camera's frame. This method allows the projection to be in a projector's view point of what the perception sensor sees.

To have the same optical properties as the projector, the virtual camera subscribes to a ROS topic with the CameraInfo message. This ensures that, when objects are projected back to the real-world, the projected objects match the object's physical shape and size.

Finally, because this virtual camera resides inside Rviz, it is able to see everything being visualized in Rviz, such as interactive markers \cite{gossow2011interactive}, navigation maps, or point clouds. For example, point clouds are seen by the virtual camera and projected in Figure \ref{fig:projection}.

\subsection{Projection Output}

Because the \texttt{rviz\_camera\_stream} package publishes the image of what the virtual camera sees, we use the image viewer from the \texttt{image\_view} ROS package to subscribe to the image topic and output to the projector through HDMI. Note that the image viewer is in full screen by enabling and using the ``Toggle full screen'' keyboard shortcut in the keyboard setting of Ubuntu's Settings software.

\subsection{Hardware Platform}\label{sec:hardware}

\begin{figure}[t!]
\centering
\includegraphics[width=0.31\columnwidth]{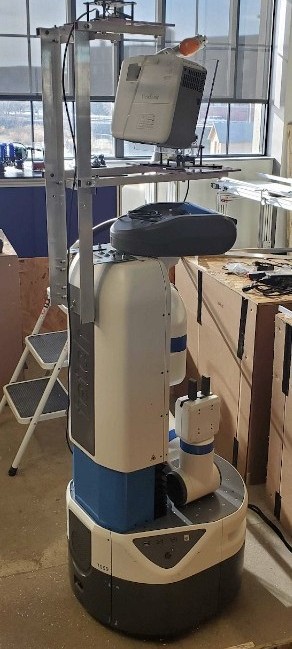} 
\caption{A sample use, where a projector is mounted onto a Fetch robot via a custom hardware structure attached to its upper back and a turret unit to pan and tilt the projector. However, the projection mapping technique is robot agnostic and the projector does not have to be attached to the robot. See Section \ref{sec:hardware} for more details.}
\label{fig:robot}
\end{figure}

As seen in Figure \ref{fig:robot}, we have demonstrated an implementation on a Fetch robot with a custom structure to mount a projector. While it would have been easier to attach the projector to the robot's head, the neck may not have had enough torque to bear it, so we chose to attach the structure to the robot's upper back. In order to pan and tilt the projector, we attached a ScorpionX MX-64 Robot Turret Kit\footnote{\url{https://www.trossenrobotics.com/p/ScorpionX-RX-64-robot-turret.aspx}}.

Despite using a Fetch robot, the projection mapping technique we describe is robot agnostic. Previously, we also applied it on an assistive robot \cite{wang2018towards}. The only requirement is that there is a transform frame for the projector so it is integrated into the transform hierarchy of a robot. In our case, we created two frames: one for the projector lens and another for the attachment point at the bottom of the projector. However, this requirement does not necessarily mean that the projector must be attached to the robot, but instead that it must be co-located with the robot in the operating environment. For example, if one would like to use projection mapping with an industrial robot arm, the projector can be placed nearby on a structure as long as its lens is pointing at what the perception sensor is targeting. By doing so, a robot arm will also not block the projection during manipulation, as careful readers may notice in the accompanying video. Otherwise, one can add a cylinder collision object to model the line of sight to mitigate blockage.

\section{Conclusion}

In this paper, we have detailed our approach to and implementation of projection mapping in robotics. Four major components were discussed: projector consideration and calibration, a virtual camera in Rviz, projection output, and a sample hardware platform.

To implement projection mapping, a roboticist can purchase an off-shelf projector, calculate a coordinate frame after installing it, calibrate its lens to get the lens intrinsics, and publish it in an encapsulated CameraInfo message. Then the roboticist would set up a virtual camera using the \texttt{rviz\_camera\_stream} Rviz plugin and subscribe to the CameraInfo message. Then the roboticist could add point cloud visualizations or any other displays (e.g., interactive markers) in Rviz and have the projector point them either manually or using a turret unit, and finally use an \texttt{image\_viewer} to output the image from the virtual camera to the projector.

\section*{Acknowledgments}
This work has been supported in part by the Office of Naval Research (N00014-18-1-2503) and the National Science Foundation (IIS-1763469). Thanks to Analog Devices (ADI) and MassRobotics for organizing the ADI Sensor Fusion Challenge\footnote{\url{https://www.mradichallenge.com/}}, providing funds to purchase the sensors.
We also thank Brian Flynn for building the hardware structure and Daniel Giger for assembling the turret unit.

\bibliographystyle{aaai}
\bibliography{bib}

\end{document}